%% file: aaai2024.tex
\documentclass[letterpaper]{article} 
\usepackage{aaai24}  
\usepackage{times}  
\usepackage{helvet}  
\usepackage{courier}  
\usepackage[hyphens]{url}  
\usepackage{graphicx} 
\usepackage{natbib}  
\usepackage{caption} 
\usepackage{algorithm}
\usepackage{algorithmic}

\usepackage{tabularx}
\usepackage{algorithm}
\usepackage{algorithmic}
\usepackage{subfigure}
\usepackage{graphicx}
\usepackage{easyReview}
\usepackage{multirow}
\usepackage{xspace}
\usepackage{amsfonts, amsmath, amssymb}
\usepackage{url}
\usepackage{amsmath}

\newcommand{\bS}{\mathbf{S}}
\newcommand{\by}{\mathbf{y}}

\newcommand{\bw}{\mathbf{w}}

\usepackage{newfloat}
\usepackage{listings}
\DeclareCaptionStyle{ruled}{labelfont=normalfont,labelsep=colon,strut=off} 
\floatstyle{ruled}
\newfloat{listing}{tb}{lst}{}
\floatname{listing}{Listing}
\title{GMP-AR: Granularity Message Passing and Adaptive Reconciliation for Temporal Hierarchy Forecasting}
\author{
Fan Zhou$^{1}$, Chen Pan$^{1}$, Lintao Ma$^{1}$, Yu Liu$^{1}$ , James Zhang$^{1}$, Jun Zhou$^{1}$, Hongyuan Mei$^{2}$, Weitao Lin$^{1}$, Zi Zhuang$^{1}$, Wenxin Ning$^{1}$, Yunhua Hu$^{1}$, Siqiao Xue$^{1}$
}
\affiliations{${^1}$Ant Group, Hangzhou China \\
$^{^2}$TTIC, Chicago USA \\ 
\{hanlian.zf, bopu.pc, siqiao.xsq\}@antgroup.com}

\usepackage{bibentry}

\begin{document}

\maketitle

\begin{abstract}
Time series forecasts of different temporal granularity are widely used in real-world applications, e.g., sales prediction in days and weeks for making different inventory plans. However, these tasks are usually solved separately without ensuring coherence, which is crucial for aligning downstream decisions. Previous works mainly focus on ensuring coherence with some straightforward methods, e.g., aggregation from the forecasts of fine granularity to the coarse ones, and allocation from the coarse granularity to the fine ones. These methods merely take the temporal hierarchical structure to maintain coherence without improving the forecasting accuracy. In this paper, we propose a novel granularity message-passing mechanism (GMP) that leverages temporal hierarchy information to improve forecasting performance and also utilizes an adaptive reconciliation (AR) strategy to maintain coherence without performance loss. Furthermore, we introduce an optimization module to achieve task-based targets while adhering to more real-world constraints. Experiments on real-world datasets demonstrate that our framework (GMP-AR) achieves superior performances on temporal hierarchical forecasting tasks compared to state-of-the-art methods. In addition, our framework has been successfully applied to a real-world task of payment traffic management in Alipay by integrating with the task-based optimization module.
\end{abstract}

\section{Introduction}
Time series forecasting is widely used in many important real-world tasks, such as supply chains management~\cite{athanasopoulos2009hierarchical, demandforecast13}, traffic flow prediction~\cite{traffic2017,li2018diffusion}.
Forecasting of different temporal granularity is essential for informed decision-making in downstream tasks. In management and operational scenarios, forecasting at different time scales is necessary to formulate plans that span from short-term to long-term. For instance, in the retail sales industry, managers need to make decisions about how much inventory to order and how to allocate it to different stores. Daily predictions can be used to plan the daily commodity distribution to each store, while weekly predictions can be used to purchase commodities in bulk. This ensures that the stores have the proper amount of inventory on hand to meet customer demand, while also minimizing the cost of carrying excess inventory.

\begin{figure}[t]
    \centering
    \includegraphics[width=0.8\linewidth]{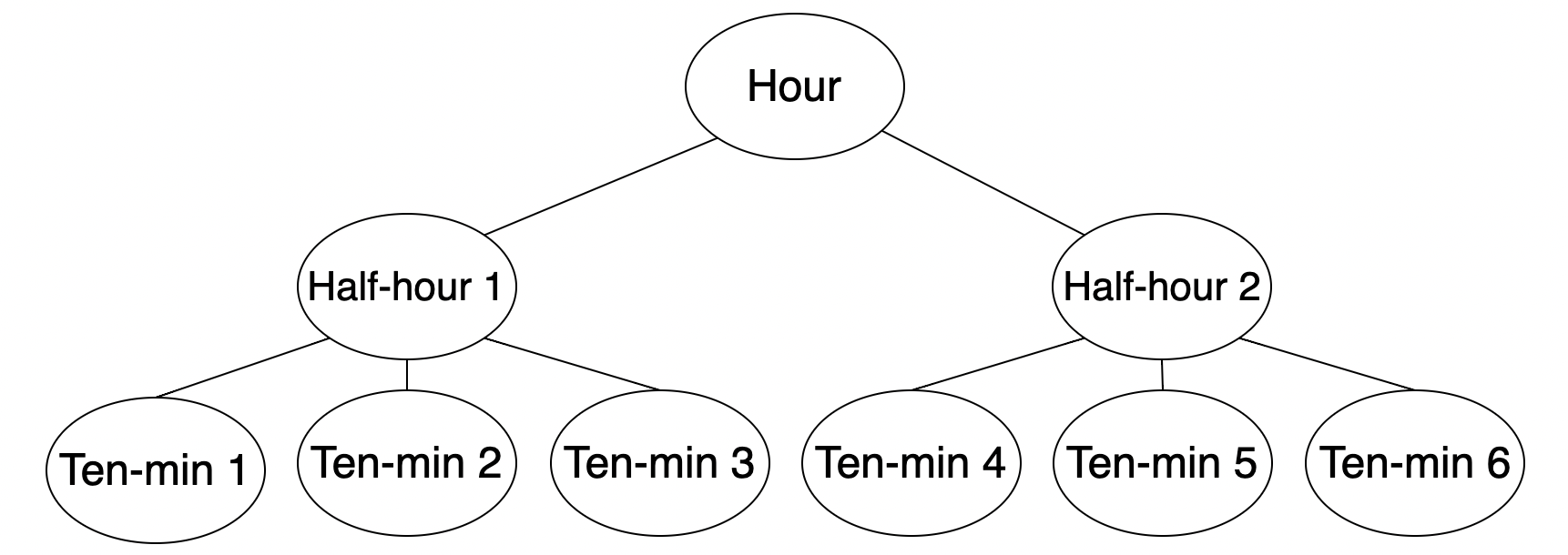}
    \caption{An example of temporal hierarchical time series (HTS) structure for ten-minute frequency as the bottom level, half-hour and hour frequencies as aggregated levels of coarser granularity.}
    \label{fig:temporal_hts}
\end{figure}

As shown in Fig.~\ref{fig:temporal_hts}, a temporal hierarchical structure is formed naturally with different levels of temporal granularity. The time series of the finest granularity makes up the bottom level, and it is aggregated into upper levels of coarser granularity. This type of time series is commonly referred to as a  \textit{temporal hierarchical time series} or temporal HTS ~\cite{thts_2017}.
Temporal HTS forecasting tasks are challenging because they require accurate prediction results for different granularity while satisfying aggregation (coherence) constraints that time series at upper levels are the aggregation of those at lower levels. 
It is difficult to produce accurate predictions for all levels with a single model because time series at each level have different granularity and exhibit their own specific dynamics, such as different trends and various seasonalities.
In addition, independent forecasts (i.e., \textit{base forecasts}) are unlikely to adhere to the coherence constraints across temporal granularity. 
The inaccurate and incoherent forecasts would result in inefficient planning of downstream tasks and pose risks to decision-making.

Previous works on temporal hierarchy forecasting can be categorized into two types: vanilla forecasting methods for temporal HTS and multivariate HTS methods applied to temporal hierarchy. 
Most temporal HTS methods~\cite{thts_2017, theodosiou2021forecasting} are statistical and theoretically explainable, but forecast performance is limited when applied to complex real-world datasets, implemented in the {\sc{thief}} package. 

The multivariate HTS methods typically follow a two-stage paradigm. In the first stage, base forecasts are generated independently for each time series in the hierarchy with statistical or deep learning models. In the second stage, base forecasts are adjusted via reconciliation to ensure coherence.

Most methods ignore the temporal hierarchical information among levels of different granularity in forecasting, one exception is COPDeepAR~\cite{rangapuram2023}, which leverages graph neural networks (GNNs) to extract structure information. 
However, GNNs introduce noise connections between nodes in the tree structure and cause performance loss.
As for reconciliation, traditional statistical methods (e.g., MinT ~\cite{wickramasuriya2019optimal}) derive coherent forecasts relying on strong statistical assumptions. The end-to-end method~\cite{rangapuram2021end} ensures coherence without strong assumptions by using a closed-form projection. However, this may reduce the forecasting performance because the adjustment scale is sometimes either too large or small and could not adapt to node values.

In this paper, we propose a framework (GMP-AR) that efficiently utilizes information among granularity at different levels of temporal hierarchy to improve the forecast performance while maintaining coherence. In summary, our contributions are as follows:
\begin{itemize}
    \item 
    We propose a granularity message passing mechanism (GMP) to enrich the information of temporal input for each node and integrate the temporal hierarchical features among different granularity to generate base forecasts.
    This is the first method that employs the temporal hierarchical structure for forecasting tasks to the best of our knowledge.
    \item We provide an adaptive reconciliation method to produce coherent results without forecasting performance loss by utilizing the node-dependent weighted optimization to precisely control the adjustment scale node by node. 
    \item We integrate our framework with an optimization module to solve real-world problems with task-based targets and realistic constraints.
    \item Experiments on real-world time series datasets demonstrate that our proposed approach achieves significant improvements over the state-of-the-art baselines and our framework combined with the optimization module has been deployed on the payment traffic management system in Alipay.
\end{itemize}


\section{Preliminary and Related Work}

In this section, we provide background on temporal hierarchical time series and introduce necessary notations here for ease of understanding.
\begin{figure}[t]
    \centering
    \includegraphics[width=0.7\linewidth]{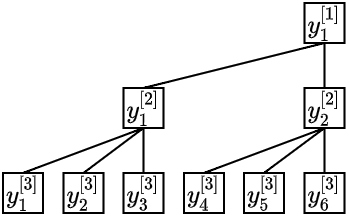}
    \caption{The notation of each node in Fig.~\ref{fig:temporal_hts} observed at timestamp $t = 6$ and $\tau  = 1$.}
    \label{fig:temporal_hier}
\end{figure}

\subsection{Preliminaries: Temporary Hierarchical Structure}
The time series is denoted by $\{y_t; 1 \le t \le T\}$ with a fixed sampling frequency ({\it base frequency}), which is of the finest granularity. The time series of the coarser granularity is aggregated by non-overlapping and regularly sampled values $y_t$ as Eq.~\eqref{agg_expression} shows. For the time series in Fig. \ref{fig:temporal_hts}, the base frequency is ten-minute. It is aggregated into a half-hour value with three ten-minute sample values, and an hourly frequency with two half-hour sample values.

The temporal HTS can be expressed as a tree structure with a linear aggregation constraint called \textit{coherence constraint}, formulated via an aggregation matrix $\mathbf{S}\in \mathbb{R}^{n \times m}$ (see Eq.~\eqref{eq_sum}), where $m = 6$ is the number of leaf nodes and $n = 9$ is the total number of nodes in the tree in Fig.~\ref{fig:temporal_hts}.
 
We consider a \textit{p-level} aggregated temporal hierarchical structure. The number of nodes at each level can be presented as $N_p = \{n_k: k  = 1, 2, \dots, p \}$, where $n_1 = 1$, and $n_p = m$. Take Fig.~\ref{fig:temporal_hts} as an example, $N_3 = \{1, 2, 6\}$. The sample values at timestamps in the time series at level $k$ can be expressed as

\begin{equation}
\label{agg_expression}
    y_{(\tau - 1)n_k+j}^{[k]} = \sum_{t=(\tau-1) m +(j-1)m_k+1} ^ {(\tau-1) m+jm_k} y_t, \,1\le j \le n_k,
\end{equation}
where $\tau \in \{1,\dots, \lfloor\frac{T}{m}\rfloor\}$ is the observation index of the
most aggregated time series (i.e., the one at the root level), and $j$ is the observation index at level $k$ in the temporal structure, and $m_k = \frac{m}{n_k}$ is the number of leaves aggregated to generate the node values at the $k$-th level. One can see that the values $\{\displaystyle{y^{[k]}_{\tau n_k}},1\le k \le p\}$ are observed at the same time point $\tau m$.

In the temporal hierarchical structure, the leaf nodes are called the \textit{bottom-level} series: $\by^{[p]}_{\tau}\in \mathbb{R}^{m}$, and the remaining nodes are termed \textit{upper-levels} series: $\{y^{[1]}_{\tau}, \dots, \by^{[p-1]}_{\tau} \}$.  The number of nodes at upper levels is $r=\sum_{i=1}^{p-1} n_i$.
Obviously, the total number of nodes $n=r+m$, and $\by_{\tau} :=[y^{[1]}_{\tau}, {\by^{[2]}_{\tau}}^{\mathsf{T}}, \dots, {\by^{[p]}_{\tau}}^{\mathsf{T}}]^{\mathsf{T}} \in  \mathbb{R}^{n} $ contains observations at time $\tau $ for all levels, which satisfies  
\begin{equation}
\label{eq_sum}
\by_{\tau }=\bS \by^{[p]}_{\tau },
\end{equation}
where $\bS\in \{0,1\}^{n \times m}$ is an aggregation matrix. 

Taking Fig.~\ref{fig:temporal_hts} as an example, the aggregation matrix 
is

\begin{equation*} 
    \bS = 
    \begin{pmatrix}
        \bS_{\text{sum}}\\
        \mathbf{I}_6
    \end{pmatrix}
    =
    \begin{pmatrix}
    \begin{array}{cccccc}
        1 & 1 & 1 &1 & 1 &1 \\
        1 & 1 & 1 &0 &0 &0\\
        0 & 0 & 0 &1 &1 & 1\\
        \hdotsfor{6}\\
        & & \mathbf{I}_6 & &\\
     \end{array} 
     \end{pmatrix}
     ,
\end{equation*}
the total number of nodes in the hierarchy is $n=1+2+6$, and the number of notes at upper-levels is $r=3$.  At each time index $\tau$, 
the coherence constraint of Eq.~\eqref{eq_sum} can be represented as ~\cite{rangapuram2021end}
\begin{equation}
\label{eq:coherent_co}
\mathbf{A}\by_{\tau}=\mathbf{0},
\end{equation}
where $\mathbf{A}:=\left (\mathbf{I}_{r}|-\bS_{\text{sum}}\right ) \in \{0,1\}^{r \times n}$, $\mathbf{0}\in \mathbb{R}^r$ is the zero vector, and $\mathbf{I}_{r} \in \mathbb{R}^{r \times r}$ is the identity matrix.  

\subsection{Related Work} 

\subsubsection{Temporal Hierarchical Forecasting.} 
The temporal hierarchical forecasting defined in~\cite{thts_2017} requires coherence constraint among temporal granularity,
and forecasting tourism data with coherence on both cross-sectional and temporal dimensions is studied in~\cite{Kourentzes2019Cross}. They rely on statistical tools for the reconciliation of base forecasts. \cite{multiscale_rnn21} is also aimed at improving the forecasting accuracy of time series with different levels of granularity. However, this work does not address coherence. 
Recently, an end-to-end model (COPDeepAR) is proposed in~\cite{rangapuram2023} that focuses on generating coherent probabilistic forecasts for time series with various levels of granularity. This is achieved by utilizing GNNs to extract inter-level information and applying a closed-form projection reconciliation method to maintain coherence. 
GNNs introduce spurious connections among all nodes but valid information actually only exists between parent and child nodes. Therefore, it fails to capture structure information because of noise connections and results in loss of forecasting performance as pointed out in \cite{SLOTH_2023}. In our works, we propose a more efficient information extraction mechanism between different granularity in the temporal hierarchy.

\section{Method}
\begin{figure*}[t]
    \centering
    \includegraphics[width=0.9\textwidth]{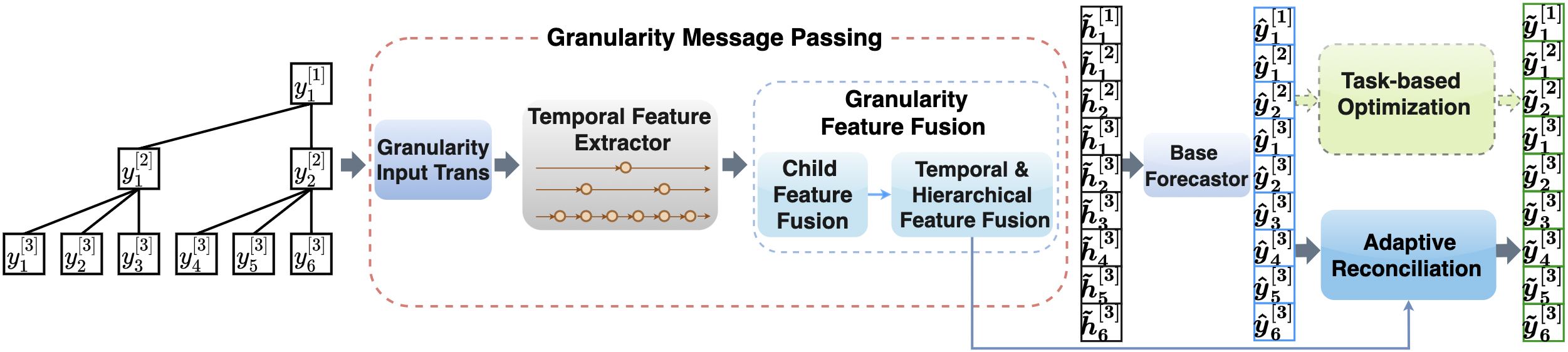}
    \caption{The architecture of GMP-AR: the red dashed box is the granularity message passing component including the granularity input transformation module, temporal feature extractor and two granularity feature fusion modules. This component generates the representation of nodes used to generate base forecasts and adaptive weights. The light green dashed box is the reconciliation module that produces the final results,
    including the adaptive reconciliation module for forecasting tasks and task-based optimization module to adapt to general real-world tasks.
    }
    \label{fig:architecture}
\end{figure*}

In this section, we introduce our framework (GMP-AR) that takes advantage of the message-passing to extract valid information among different granularity in the temporal hierarchy to improve forecasting performance and achieve adaptive reconciliation to maintain coherence.
Finally, we integrate an efficient optimization module with our framework to solve real-world problems with task-based targets and realistic constraints. The overall architecture is shown in Fig.~\ref{fig:architecture}, which consists of three main components:
\begin{itemize}
    \item {\it Granularity message passing} module: This module leverages temporal hierarchy information between different granularity to generate base forecasts over the prediction horizon across all levels and adaptive weights;
    \item {\it Adaptive reconciliation} module: This module utilizes adaptive weights combined with projection reconciliation to produce node-dependent adjustment to maintain coherency in an end-to-end way;
    \item {\it `Plug-and-play' optimization} module: This module adapts to real-world problems with task-based targets and realistic constraints.
\end{itemize}

\subsection{Granularity Message Passing}
In this section, we introduce our granularity message-passing module that incorporates both temporal and granularity hierarchical information to generate the base forecasts and the weights for adaptive reconciliation. This module consists of four sub-modules: granularity input transformation, temporal feature extractor, and two granularity feature fusion modules.

\subsubsection{Granularity Input Transformation}
\begin{figure}
    \centering
    \includegraphics[width=0.85\linewidth]{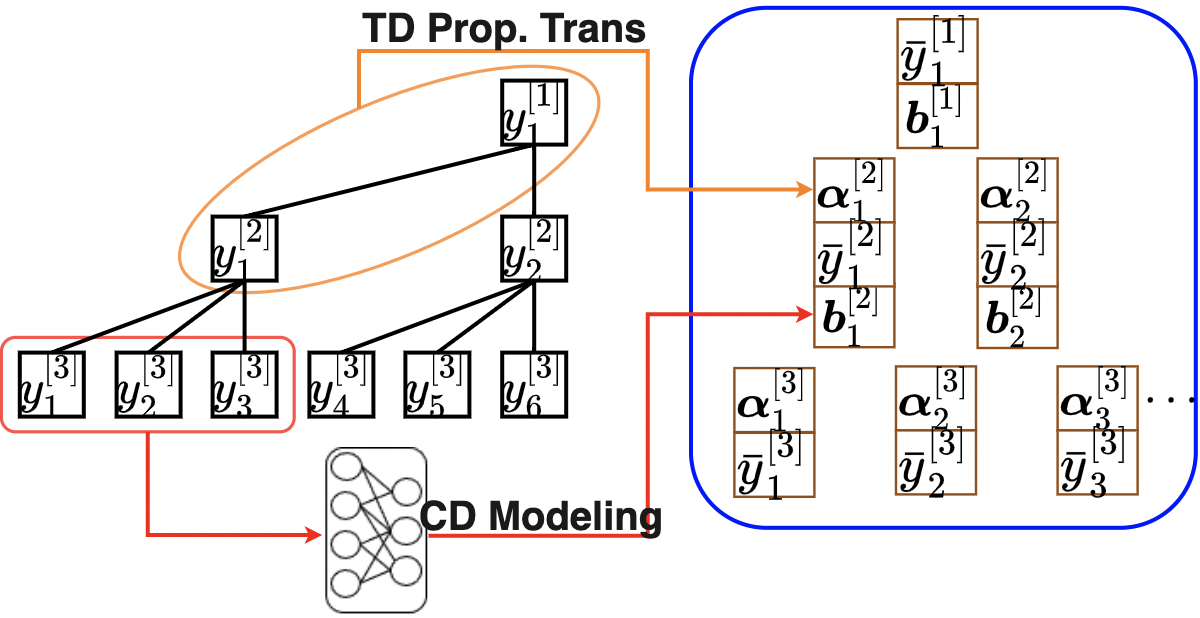}
    \caption{The process of granularity input transformation: the orange cycle is the top-down proportion transformation module that produces the disaggregation proportions for child nodes, the red rectangle is the child distribution modeling module which extracts valid information of finer granularity for parent nodes; the blue box concatenate these processed inputs with normalized value and put them into temporal feature extractor to extract dynamic patterns.}
    \label{fig:input_trans}
\end{figure}

This module integrates the input information of different granularity in the hierarchy into temporal inputs by \textit{top-down proportion transformation} and \textit{child distribution modeling} as in Fig.~\ref{fig:input_trans}.

\textit{Top-Down Proportion Transformation.}
Similar to the top-down proportion models in TDProb~\cite{das2022deep},
we transform the inputs $\{\by_1, \dots, \by_{\tau}\}$ as the fractions/proportions according to root time-series which is disaggregated
into its child time series as in Fig.~\ref{fig:input_trans}. The detail is as follows
\begin{align}
    a^{[k]}_{\tau,j} = \frac{y^{[k]}_{(\tau-1)n_k  + j}}{y^{[1]}_{\tau}}, \quad \tau  \in \left\{1, 2, \dots, \lfloor\frac{T}{m}\rfloor \right\},
\end{align}
where $a^{[k]}_{\tau,j}, 1\le j \le n_k$ is the fraction input at time $\tau $. Then we sort the fraction inputs from the root node to the current node to build the top-down proportion inputs $\boldsymbol{\alpha}_{\tau,j}^{[k]} = [a^{[2]}_{\tau, l_2}, \dots,  a^{[k]}_{\tau,j}],k\ge 2$, where $l_2$ means the node index at the second level on the path from the root to the current node, and {$l_i = \lceil l_{i+1} / \frac{n_{i+1}}{n_{i}} \rceil$}.  We use fraction inputs starting from the second level, as the fraction for the root level is always equal to $1$ and does not provide any meaningful information for the temporal extraction module.

The intuition behind this approach is as follows: 1) As pointed out in \cite{das2022deep}, the disaggregation proportions of nodes at the bottom level are more predictable compared to raw values of time series; 2) Incorporating the information of nodes at the top level with coarser granularity into the leaf nodes of finer granularity at the bottom level to enhance temporal stability, which is based on the analysis that the seasonality of the top level is clearer and the evolutionary dynamic is 'smoother' \cite{demandforecast13, taieb2021hierarchical}.

\textit{Child Distribution Modeling.}
This sub-module integrates the disaggregation proportion distribution of the child nodes with finer granularity into the temporal input of the parent node. This enriches the finer granularity information of the parent node, which help capture dynamic patterns more effectively for parent node with coarser granularity, improving the overall forecasting accuracy.

Our approach utilizes deep models to extract the distribution of child nodes' disaggregation proportions for each parent node.  Specifically, 
we assume that each node at the same level $k$ has the same number of children $m_k$, so the hierarchical structure is fixed for each upper level. Given the fixed structure, MLP~\cite{gardner1998artificial} is an efficient tool to extract features, 
\begin{align}
    \mathbf{b}^{[k]}_{\tau, j} = \text{MLP}\left([a^{[k+1]}_{\tau,(j-1)m_{k}+l}: 1\le l \le m_{k}]\right),\,k < p,
\end{align}
where $\mathbf{b}^{[k]}_{\tau, j}$ represents the hidden features of child node's distribution for the parent node $y^{[k]}_{(\tau - 1) n_k  + j}$ and $\{a^{[k+1]}_{\tau,(j-1)m_{k} + l}: 1\le l \le m_{k}\}$ is the proportion input of its child set.

\subsubsection{Temporal Feature Extraction.}
This module extracts temporal features for nodes at the same level (same granularity) as follows:
\begin{align*}
    h^{[k]}_{\tau,j} &= \text{UPDATE}^{[k]}\left(h^{[k]}_{\tau,j-1,}, [\boldsymbol{\alpha}^{[k]}_{\tau,j}, \mathbf{b}^{[k]}_{\tau,j}, \Bar{y}^{[k]}_{\tau,j}]; \theta^{[k]}\right), \\
    \mathbf{H}_{\tau } &= [h^{[1]}_{\tau,1 }, h^{[2]}_{\tau,1}, \dots, h^{[p]}_{\tau, m}] , 
\end{align*}
where $\boldsymbol{\alpha}^{[k]}_{\tau,j}$ is the top-down proportion inputs of node \textit{j} at level $k$ and time point $\tau $, $\mathbf{b}^{[k]}_{\tau,j}$ is the distribution feature of the child set, and $\Bar{y}^{[k]}_{\tau,j}$ is the scaled value of the sample at the same level over the whole past observations as follows:
\begin{align}
    \Bar{y}^{[k]}_{\tau,j} = \frac{{y}^{[k]}_{(\tau-1)n_k + j}}{\max\{y^{[k]}_{1}, \dots, {y}^{[k]}_{\tau n_k}\}},\, 1\le j \le n_k.
\end{align}
Since the root node has no ancestors and the leaf nodes have no child, we remove $\boldsymbol{\alpha}_\tau^{[1]}$ for the coarsest granularity and $\mathbf{b}_{\tau,j}^{[p]}$ for the finest granularity. 
The module $\text{UPDATE}^{[k]}$ is a temporal pattern extraction function at level $k$, and $\theta^{[k]}$ is the parameters of the function. Any type of recurrent-type neural networks can be adopted as the temporal feature extraction
module, such as the RNN variants, TCN~\cite{bai2018empirical},  and
NBeats~\cite{oreshkin2019n}. In our experiments, we use GRU~\cite{chung2014empirical} for simplicity.

It is worth emphasizing that temporal HTS differs from multivariate HTS in that different levels do not share the same temporal features. In multivariate HTS, all nodes are of the same temporal granularity and may have similar trends or seasonality.  However, in temporal HTS, nodes at different levels have different granularity and different context lengths. Nodes at the top levels,  which have coarser granularity, have smaller context lengths~\cite{rangapuram2023}.
In other words, different levels have their own dynamic patterns (trend, seasonality, etc.), which requires different specific parameters $\theta^{[k]}$ of temporal updating function for each level.

\subsubsection{Child Granularity Feature Fusion}
This module integrates temporal features from the finer granularity of child nodes to their parent at top levels, enhancing the framework's ability to adapt to the variation of dynamic patterns, such as sudden trend changes {due to some} special events, e.g., Double 11, black Friday. This because the value aggregation relationship between the parent node and child nodes determines the changes in dynamic patterns at the finer granularity would influence the ones at the coarser granularity.

Unlike directly modeling the disaggregation proportion distribution across child nodes, this module extracts valid temporal features from child nodes, which may have different contributions to their parent.  CNN and attention mechanism~\cite{vaswani2017attention} are effective tools to extract valid patterns from various features. However, we assume that all nodes at the same level have the same number of children, which indicates the tree structure at the same level is fixed.  For a fixed structure, CNN is more appropriate because it has lower computation cost (see Appendix E) and better performance as demonstrated in the ablation study (see Appendix F). The detail is as follows:
\begin{align*}
    \hat{h}^{[k]}_{\tau, j} &= \text{Conv}^{[k]}\left([h^{[k+1]}_{\tau,(j-1)m_k + l}: 1 \le l \le m_{k}]\right)  \\
    &= \sum_{l=1}^{m_{k}} w^{[k]}_l h^{[k+1]}_{\tau, (j-1)m_k + l}\;,
\end{align*}
where $\text{Conv}^{[k]}$ is the CNN kernel for level $k$, and the $w^{[k]}_l$ is the kernel weight of child $l$ at level $k$.

\subsubsection{Temporal and Hierarchical Granularity Feature Fusion.}
This module integrates the temporal and hierarchical features among granularity for each node to enrich the dynamic and structure information of node representation, which is used to generate the base forecasts and the weights for adaptive reconciliation.

We adopt CNNs to extract valid information across the temporal and granularity hierarchy domains as follows:
\begin{align*}
    \Tilde{h}^{[k]}_{\tau, j} &= \text{Conv}(\hat{\mathbf{H}}^{[k]}_{\tau, j}; \theta), \\
    {\mathbf{\hat{H}}^{[k]}_{\tau, j}} 
    &= \begin{pmatrix}
        {h}^{[k]}_{1, j} & \hat{h}^{[k]}_{1, j}\\
        \vdots & \vdots \\
        {h}^{[k]}_{\tau,j} & \hat{h}^{[k]}_{\tau, j}
    \end{pmatrix},
\end{align*}
where all nodes share the same extraction parameters, $\Tilde{h}^{[k]}_{\tau, j}$ are used as inputs of MLPs (the module can be replaced with other neural layers, such as seq2seq~\cite{sutskever2014sequence}, transformers~\cite{vaswani2017attention}, and CNNs) to generate the base forecasts $\hat{\by}_\tau$ and reconciliation weights $\mathbf{w}$.

\subsection{Adaptive Reconciliation}
In this section, we introduce our adaptive reconciliation module by utilizing node-dependent weights that incorporate temporal and hierarchical information. These weights are used to control the adjustment scale node by node in reconciliation, improving the prediction performance and ensuring coherence.

Under the assumption that base forecasts are highly accurate, the closed-form projection reconciliation only minimizes the distance between the coherent result with the base forecast~\cite{rangapuram2021end}. However, the optimization target here is to minimize the mean distance among all nodes, which may result in a performance loss for several reasons. 
1) The scale of values among different nodes can be diverse, and mean distance minimization may cause a relatively large adjustment to the node with a small value, while only slight adjusting to the nodes with large values. Since the values at higher levels of the hierarchy have a coarser granularity that are larger than those at lower levels, the reconciliation mainly provides efficient adjustments to the upper level values. 
2) Since different nodes have distinct levels of forecasting accuracy,  the reconciliation adjustment scale should also be insignificant for nodes with high accuracy. In order to achieve adaptive adjustment node by node, node-dependent weights are imposed to the optimization target to resolve these issues as follows
\begin{equation}
    \begin{aligned}
    \label{eq:weight_recon}
     \Tilde{\by}_\tau & = \mathop{\arg\min}_{\by \in \mathbb{R}^n} \|\bw(\by - \hat{\by}_\tau) \|_2 \\
      \text{s.t.} &\quad \mathbf{A}{\by} = \mathbf{0},
    \end{aligned}
\end{equation}
where $\bw \in \mathbb{R}^{n \times n}$ is a diagonal matrix {with weights on the diagonal}. Then we apply the method of Lagrange multiplier to derive the closed-form solution (see the proof in Appendix A):
\begin{equation}\label{eq:analytic_sol}
    \Tilde{\by}_{\tau } = \mathbf{M} \hat{\by}_{\tau },
\end{equation}
where $\mathbf{M} = (\mathbf{I} - \tilde{\bw}^{-1}\mathbf{A}^\mathsf{T}(\mathbf{A}\tilde{\bw}^{-1}\mathbf{A}^\mathsf{T})^{-1}\mathbf{A}) \hat{\by}_\tau$, $\tilde{\bw} = {\bw^\mathsf{T}}{\bw}$.

The node-dependent weights could be hyperparameters for simpler structures or realistic tasks, which is explainable and easy to implement.  However, it requires prior {domain} knowledge of each scenario to set {proper} weights, which causes inflexibility to new scenarios and the improvement of performance is limited.
Our framework utilizes temporal and hierarchical granularity to generate the node-dependent weights, which has superior performance improvement and adaptability than human settings as shown in Appendix F.

\subsection{Task-based Optimization in Real-world Scenarios}
This module introduces an optimization module, which can be applied in our forecasting framework (GMP-AR) to solve real-world problems with task-based targets and realistic constraints. The detailed formulation is as follows:
\begin{equation}
    \label{eq:recon_task}
    \begin{aligned}
      &\mathcal{J}(\hat{\by}) = \mathop{\min}_{\by}{f(\hat{\by}, \by)}\\
    \text{s.t.}\quad 
    &\begin{cases}
    \mathbf{A}\by = \mathbf{0}, e_j = 0, \; j=1, \dots, n_{\text{eq}},\\
    g_i(\by, \hat{\by}) \le 0, \; i=1, \dots, n_{\text{ineq}}.
    \end{cases}
     \end{aligned}
\end{equation}
The target functional $f$ can take various forms of functions besides Euclidean distance (such as cosine similarity and $L^d$-norm ($d\ge 1$)){. The} $e_j$ terms represent task-based equality {constraints} other than coherence constraint{. The} $n_{\text{eq}}$ term is the number of equality constraints, the $g_i$ term is an inequality constraint, and $n_{\text{ineq}}$ term is the number of task-based inequality constraints. 
As shown in Eq.~\eqref{eq:recon_task}, our task is transformed into an optimization task, where legacy methods such as L-BFGS or Powell~\cite{num_optim} can be applied.  Many useful optimization methods have been encapsulated in \textit{Scipy.optmize}, which is used in our experiment in Alipay with finance targets and supervised constraints.

\section{Experiments}
In this section, we empirically evaluate our proposed method on public time series datasets (details are in Appendix B) and 
representative SOTA methods (details are shown in Appendix C).
We then present the successful application of our method with the task-based optimization module in the real-world payment traffic management system of Alipay.  The experiment setup is described in Appendix G and the implementation code is included in the supplementary files.

\begin{table*}[t]
    \centering
    \resizebox{1.\linewidth}{!}
    {
    \begin{tabularx}{1.08\textwidth}{l|cc|cc|cc}
        \hline
         \textbf{Dataset} & \multicolumn{2}{c}{Electricity}  & \multicolumn{2}{c}{Traffic} & \multicolumn{2}{c}{Exchange Rate}  \\ \hline
         \textbf{Metrics} & b-MAPE & MAPE & b-MAPE& MAPE & b-MAPE & MAPE  \\ \hline
         THIEF-ARIMA-BU &0.1895 & \textbf{0.1319} & 0.7501 & 0.5518 &  0.0095 & 0.0092\\ 
         THIEF-ETS-BU  &0.3712 &  0.3111 & 0.8815&  0.8600 &0.0100 & 
        0.0096)\\
         THIEF-THETA-BU  & 0.2071 &  0.1434 & 0.7064 & 0.7065 & 0.0102&   0.0099\\
         THIEF-ARIMA-OLS  & 0.1896 & 0.1319& 0.8293 & 0.5711 &  0.0118 & 0.0113\\
         THIEF-ETS-OLS  & 0.2208& 0.1438 & 1.0477 & 0.6871 & 0.0133& 0.0102 \\ 
         THIEF-THETA-OLS  &0.2256 & 0.1520 & 1.6886& 0.8952 & 0.0099 & 0.0094 \\
         THIEF-ARIMA-MSE  & 0.1906 & 0.1333 &  0.7000 & 0.5141 & 0.0097 & 0.0093\\
         THIEF-ETS-MSE  &0.2142 &  0.1473& 0.7315& 0.6872& 0.0103& 0.0100\\
         THIEF-THETA-MSE  & 0.2036 & 0.1373 &  0.9467 &  0.7035 & 0.0094& 0.0090\\ \hline
         DeepAR-BU & 0.4168(0.0182) & 0.4210(0.0154) &1.3439(0.0172) & 0.8885(0.0219) & 0.0138(0.0027) &  0.0136(0.0030) \\ 
         NBeats-BU & 0.2683(0.0383)& 0.3000(0.0318) & 0.5601(0.0570)& 0.4292(0.0377)&0.0196(0.0062)& 0.0192(0.0064)\\ 
         Autoformer-BU & 0.4059(0.1022) & 0.4049(0.0859) & 1.6798(0.2757) & 0.9505(0.1362)&0.0092(0.0006) &0.0087(0.0006) \\
         DeepAR-Proj & 0.4167(0.0042) & 0.3889(0.0023) & 3.3717(0.0290) &1.6043(0.0138) & 0.0114(0.0019) & 0.0111(0.0019)\\ 
         NBeats-Proj & 0.2432(0.0173)& 0.2799(0.0203) & 0.7422(0.0676) & 0.5308(0.0297) &0.0211(0.0072) & 0.0207(0.0075)\\ 
         Autoformer-Proj &0.2754(0.0047) & 0.3024(0.0064) & 1.1376(0.0758)& 0.6768(0.0382) & 0.0111(0.0007) & 0.0107(0.0007)\\ 
         DeepAR-TDProb & 0.1761(0.0055) & 0.2370(0.0035) & 1.0536(0.0296) &  0.6455(0.0126) & 0.0096(0.0008) & 0.0094(0.0008) \\ 
         NBeats-TDProb &0.1830(0.0081) & 0.2395(0.0098) & 1.0744(0.0339)&0.6662(0.0142) &0.0161(0.0060) & 0.0159(0.0061)\\ 
         Autoformer-TDProb & 0.2052(0.0243) & 0.2704(0.0268) & 0.9894(0.0395)& 0.6244(0.0198) & 0.0120(0.0035) &0.0118(0.0035)  \\ \hline
         HierE2E & 0.2346(0.0148) & 0.2980(0.0139)& 0.4728(0.0180) & 0.4768(0.0128) & 0.0145(0.0017) & 0.0136(0.0016)\\ 
         SHARQ & 0.2101(0.0058) & 0.2730(0.0051) &0.5041(0.0086) & 0.5041(0.0128) & 0.0156(0.0043)& 0.0138(0.0033)\\ 
         COPDeepAR &  0.6088(0.0531) & 0.7309(0.0542) & 0.8345(0.0058)& 0.8202(0.0060) & 0.1247(0.0274)& 0.1246(0.0275) \\
         SLOTH &  0.1765(0.0022)& 0.2424(0.0027) & 0.4621(0.0026)& 0.4616(0.0019) & 0.0114(0.0003)& 0.0110(0.0003)
         \\ \hline
         
         GMP-BU &0.1711(0.0074) & 0.2288(0.0079) & 0.4484(0.021) & 0.3932(0.0145)& 0.0107(0.0014)& 0.0102(0.0015) \\ 
         GMP-TDProb & 0.1731(0.0026) & 0.2331(0.0044) & 0.9989(0.0301) & 0.6047(0.0112)& 0.0088(0.0009)& 0.0085(0.0009)\\ 
         GMP-Proj &0.1937(0.0143) & 0.2481(0.0131) & 0.4868(0.0243) & 0.4083(0.0094)  &0.0089(0.0024) & 0.0086(0.0025) \\ 
         GMP-AR & \textbf{0.1499(0.0021)} & 0.2158(0.003) & \textbf{0.4289(0.0202)}& \textbf{0.3798(0.0066)} &\textbf{0.0085(0.0021)} & \textbf{0.0082(0.0021)}\\ \hline
    \end{tabularx}
    }
    \caption{B-MAPE (bottom MAPE) and MAPE metric values over five independent runs for baselines such as traditional reconciliation methods, deep learning methods, and popular multivariate HTS methods,  as well as our approach. The values in brackets are the variances over five runs.}
    \label{tab:result}
\end{table*}

\subsection{Result Analysis}
In this section, we {evaluate} the {forecasting} performance of our proposed method on three public datasets.  We use the mean absolute percentage error to measure the performance of each method, both at the bottom level (b-MAPE) that has the finest granularity and across all nodes (MAPE).

The experimental results are {shown} in Table~\ref{tab:result}.  The top section shows the results of the traditional statistic method, Thief; The second section {shows} results from deep neural networks methods with bottom-up (BU),  closed-formed projection (Proj)~\cite{rangapuram2021end} and top down proportion (TDProb)~\cite{das2022deep} reconciliation; The third section {shows the results of} popular multivariate hierarchical time series forecasting methods (HierE2E, SHARQ, COPDeepAR, SLOTH); The bottom section {shows the results of} our forecasting approach (GMP) and the combination of our forecasting mechanism with popular reconciliation methods (BU, Proj, TDProb), as well as our adaptive reconciliation with weighted projection (AR) method.

For the bottom-level metrics (b-MAPE), DeepAR~\cite{salinas2020deepar} with TDProb reconciliation performs the best on the Electricity and Exchange Rate datasets, and NBeats with BU reconciliation performs the best on the Traffic datasets. In general, the performance of statistical methods in Thief is more stable {than the} deep learning models, but deep models achieve the best performances across all baselines. 
One can observe that models of best performance on b-MAPE do not necessarily perform as well on MAPE, which is caused by different level contributions to the overall performance. For example,  the temporal patterns and scales of different levels can be diverse.
For overall metrics, statistic methods in Thief outperform other baselines and our methods on Electricity datasets, and multivariate HTS methods have superior performance than deep learning methods. This  because HTS methods take all nodes into consideration while modeling.

Our proposed approach, GMP-AR, delivers an average performance increase of over $2\% \sim 3\%$ compared to other methods on both b-MAPE and MAPE metrics in most scenarios on the Electricity and Traffic datasets. Specifically, GMP with AR Projection achieves the best performance for b-MAPE among all models on all three datasets, and it also performs the best on MAPE on Traffic and Exchange Rate datasets.
In addition to adaptive reconciliation, we also combine our forecasting mechanism (GMP) with the aforementioned popular reconciliation methods (BU, Proj TDProb). These combinations also achieve higher accuracy than the baselines.  
We also assess the performance across all levels (Appendix D) and running time (Appendix E), and the ablation study for each component in the GMP forecasting framework (Appendix F). 

In conclusion, our GMP-AR mechanism achieves the best forecasting accuracy by utilizing both temporal and granularity hierarchical information. The adaptive reconciliation produces coherent results while also improving the forecasting performance.

\subsection{Alipay Payment Traffic Management}
\begin{figure}[t]
    \centering
    \includegraphics[width=1.\linewidth]{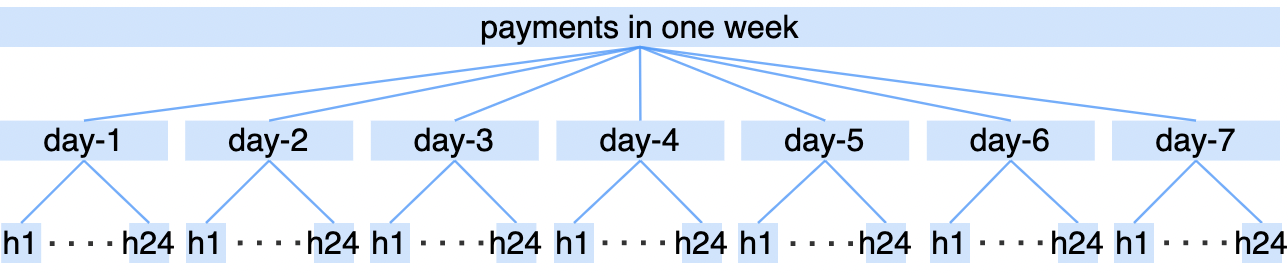}
    \caption{The temporal hierarchical structure of mobile payment service of Alipay, with weekly traffic forecast as the root node, daily forecast as the second level, and hourly forecast as leaf nodes ($N_3 = \{1, 7, 168\}).$}
    \label{fig:alipay}
\end{figure}

Alipay is a world-leading third-party online payment service platform that provides billions of users with online and mobile payment services. These services support amount to billions of transactions and trillions of dollars daily, which has been widely recognized as a significant contribution to the orderly development and digital upgrading of society and economy. To ensure a stable and reliable payment experience for users, it is necessary to forecast future payment traffic in advance to ensure that the massive {number} of payment transactions and amounts can be effectively supported at the infrastructure and operational levels. 
In the temporal dimension, the manager needs to reschedule the underlying system resource under different granularity, i.e., hourly, daily, and longer-term like weekly. This way could help the system to support billions of payments efficiently and energy-savingly. Therefore, it is necessary to forecast payment traffic at hourly, daily, and multi-day granularity. This creates a natural three-level tree structure (see Fig.~\ref{fig:alipay}). In {this} temporal hierarchical structure, the coherence of different granularity forecasts needs to be satisfied.

The real-world task of hierarchical time series (HTS) forecasting differs from theoretical temporal HTS forecasting for its more task-based targets and realistic constraints, which are listed as follows: 1) Target - not only must forecast accuracy be considered but also the trend of the forecast, which is vital for business corporations that interact with external banks. 2) Constraint - the root node, i.e., the base daily forecast, cannot be altered due to the historical requirements of the regulatory authority. 3) Constraint - the child nodes, i.e., the scale of adjustment must not be more than 20 percent due to the risk regulations set by the authority.  These requirements can be formulated as  
\begin{equation}
    \label{eq:alipay}
    \begin{aligned}
      \mathcal{J}(\hat{\by}) & = \mathop{\arg\min}_{\by}{\|\hat{\by} - \by\|_2 + \beta\left(1 -\frac{\hat{\by} \by}{\|\hat{\by}\| \|\by\|} \right)}\\
    \text{s.t.}\quad &
    \begin{cases}
    \mathbf{A}\by = \mathbf{0}, \\
    \by[0] -\hat{\by}[0]= 0,\\
    abs(\by -\hat{\by}) - 0.2\hat{\by} \le \mathbf{0} ,
    \end{cases}
     \end{aligned}
\end{equation}
where we append cosine similarity with a hyperparameter multiplier $\beta$ to optimization targets to improve trend forecasting performance, and also take additional two real-world constraints into consideration.
\begin{figure}[t]
    \centering
    \includegraphics[width=0.8\linewidth]{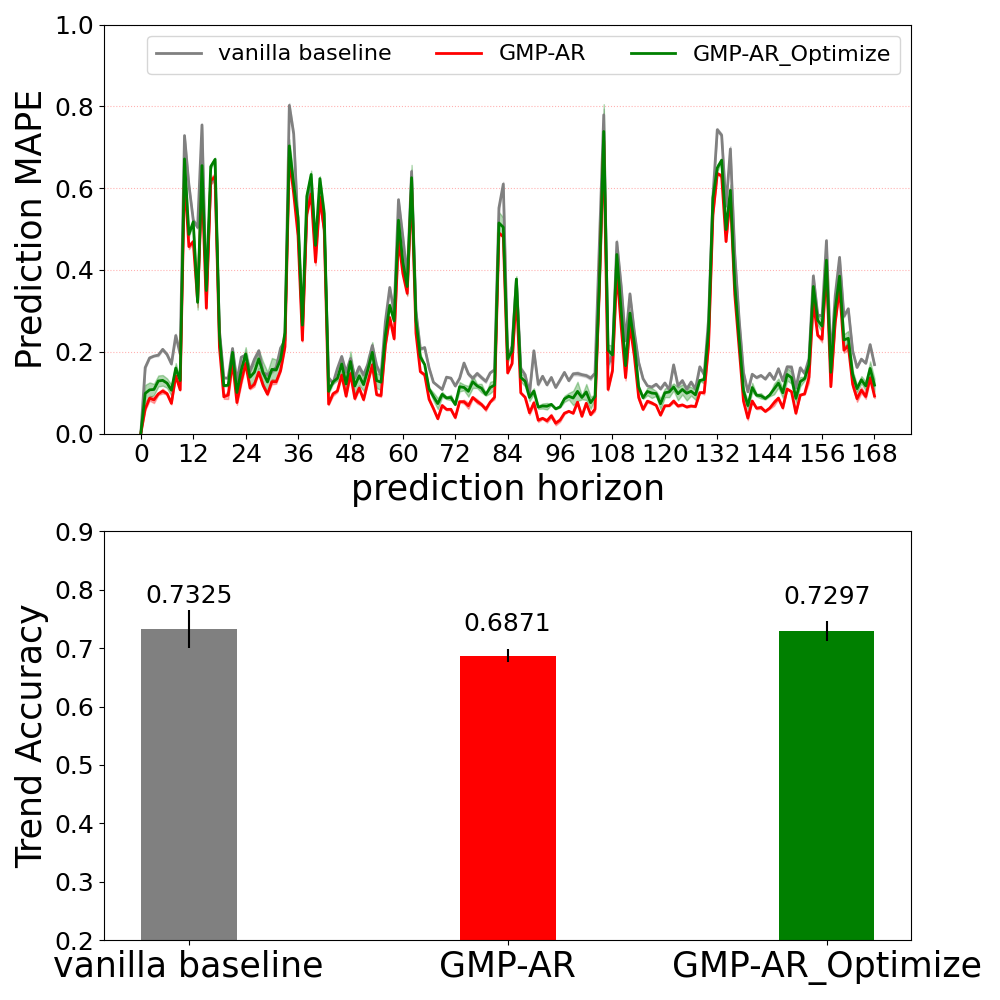}
    \caption{Results of five independent runs on Alipay scenarios of three methods, i.e., DeepAR (online baseline), GMP-AR, and GMP-AR with optimization module (lower MAPE and higher trend accuracy are better). Our GMP-AR with optimization module improves MAPE by 3.8\% over baselines without performance loss on trend accuracy.}
    \label{fig:alipay_result}
\end{figure}


{The results} in Fig.~\ref{fig:alipay_result} show that vanilla GMP-AR performs the best on prediction accuracy by 5\%-7\% without considering trend targets and the other two task-based constraints compared to the online baseline model, DeepAR. However, the trend accuracy is lower by 5\%. Our GMP-AR combined with the optimization module achieves 3\%-5\% improvements on MAPE and remains similar trend accuracy as the baseline. {In} other words, this method achieves a compromise between forecast and trend accuracy. Besides, the results of the optimization module adhere to all task-based constraints, while vanilla GMP-AR only satisfies the coherence constraint. 

\section{Conclusion}
In this paper, we introduced a novel structure-learning framework for temporal hierarchical time series (GMP-AR).
{Our framework incorporated temporal and hierarchical information with a message passing mechanism} to improve the performance of base forecasts, which is
{consisted of several modules, i.e., the granularity input transformation module that enriches the information of the temporal input of each node and the other two granularity feature fusion modules integrating the temporal and hierarchical features. }
For reconciliation, we proposed an adaptive reconciliation method to ensure coherence without any loss in forecasting performance by utilizing the node-dependent weighted optimization target to precisely control the adjustment scale for each node. 
We also provided a `plug-and-play' optimization module that incorporates task-based targets and realistic constraints to solve real-world temporal HTS problems.  
We conducted extensive empirical evaluations on real-world datasets to validate our method, demonstrating the competitiveness of our approach compared with other state-of-the-art methods. Furthermore, our ablation studies verify the efficacy of each designed component.
Our framework combining with the optimization module has been applied in the payment traffic management system in Alipay successfully.

\clearpage
\bibliography{aaai24}

\clearpage
\section{Appendix}
\input{latex/appendix}

\end{document}

%% file: latex/appendix.tex
\section{Appendix A: Proof of Adaptive Reconciliation}
\label{app:proof_ada_recon}

We derive the analytic expression~\eqref{eq:analytic_sol} here using Lagrange multiplier:
First, through introducing Lagrange multiplier $\lambda \in \mathbf{R}^n$, the original problem \eqref{eq:weight_recon} is transformed into following optimization problem without constraints:
\begin{equation*}
    \begin{aligned}
    &\,\mathop{\arg\min}_\by \frac{1}{2}\|\mathbf{w}(\by - \hat{\by}_\tau)\|_2^2+ \lambda^{\mathsf{T}}\mathbf{A} \by\\
    = &\mathop{\arg\min}_\by \frac{1}{2} \by^{\mathsf{T}} \mathbf{w}^{\mathsf{T}}\mathbf{w}\by - \hat{\by}_\tau^{\mathsf{T}}\mathbf{w}^{\mathsf{T}}\mathbf{w} \by + \lambda^{\mathsf{T}}\mathbf{A} \by.
    \end{aligned}
\end{equation*}
Then by taking derivatives with respect to $\by$ and $\lambda$, and setting them to be zero (vector), we obtain
\begin{equation*}
    \begin{cases}
    \tilde{\mathbf{w}}\by - \tilde{\mathbf{w}}\hat{\by}_\tau + \mathbf{A}^{\mathsf{T}} \lambda  &= \mathbf{0},\\
    \mathbf{A}\by &= \mathbf{0},
    \end{cases}
\end{equation*}
where $\tilde{\mathbf{w}} := \mathbf{w}^{\mathsf{T}}\mathbf{w}$ is also a diagonal matrix. Solving the above system of linear algebra equations, we then obtain
\begin{equation*}
\begin{aligned}
    \by^{*} &= \tilde{\mathbf{w}}^{-1} \left(\tilde{\mathbf{w}}\hat{\by}_\tau - \mathbf{A}^{\mathsf{T}}\lambda \right),\\
    \lambda &= (\mathbf{A}\tilde{\mathbf{w}}^{-1}\mathbf{A}^{\mathsf{T}})^{-1}\mathbf{A}\hat{\by},
\end{aligned}
\end{equation*}
and finally, we have
\begin{equation*}
    \by^* = \left( \mathbf{I} - \tilde{\mathbf{w}}^{-1}\mathbf{A}^{\mathsf{T}}\left(\mathbf{A} \tilde{\mathbf{w}}^{-1}\mathbf{A}^{\mathsf{T}}\right)^{-1}\mathbf{A}\right) \hat{\by}_\tau.
\end{equation*}

\section{Appendix B: Datasets Details}
\label{app:datasets}
\begin{table}[t]
    \centering
    \resizebox{0.95\linewidth}{!}{
    \begin{tabular}{|c|c|c|c|c|c|c}
        \hline
         Dataset & levels & nodes & structure & bottom freq & root freq \\ \hline
         Traffic & 3 & 28 & 1, 3, 24 & 1H & 1D\\ \hline
         Electricity & 3 & 28 & 1, 4, 24 & 1H & 1D \\ \hline
         Exchange Rate & 2 & 6 & 1, 5 & 1D & 5D \\ \hline
    \end{tabular}
    }
    \caption{Dataset statistics. The `structure' column shows the number of nodes at each level from top to bottom.  In frequency columns (`bottom freq' and 'root freq'), 1D means one day, 1H means one hour, and 1W means one week.}
    \label{tab:dataset_statisitic}
    \vspace{-3mm}
\end{table}
We take three publicly available datasets to conduct experiments, and the details of statistics are shown in Table~\ref{tab:dataset_statisitic}.
\begin{itemize}
    \item Traffic\footnote{http://pems.dot.ca.gov}: the road occupancy ratios (between 0 and 1) measured by different sensors on San Francisco Bay area freeways collected from the California Department of Transportation. We take a collection of 48 months (2015-2016) hourly data with 862 sensors. The temporal structure is shown in Table \ref{tab:dataset_statisitic}.
    \item Exchange Rate: the daily exchange rates of eight foreign countries including Australia, British, Canada, Switzerland, China, Japan, New Zealand and Singapore ranging from 1990 to 2016. It is worth noting that the exchange rate is not available on weekends, so we aggregate 5 weekdays as a week value.
    \item Electricity\footnote{https://archive.ics.uci.edu/ml/datasets/ElectricityLoadDiagrams20112014}: the electricity consumption dataset, which was recorded every 15 minutes from 2011 to 2014. We eliminate  records in 2011 because of missing data. Finally, we extract the electricity consumption of 321 clients from 2012 to 2014 and we convert the data to reflect hourly consumption. We construct a 3-level temporal hierarchical structure, with details shown in Table \ref{tab:dataset_statisitic}.
\end{itemize}

\section{Appendix C: Baselines}
\label{app:baselines}
We compare our method with the following prediction models and reconciliation approaches:
\begin{itemize}
    \item Traditional statistical methods: these methods explicitly incorporate temporal hierarchies forecast R package, i.e., {\sc{Thief}} \cite{thts_2017}. For Thief, we adopt the combination of three base prediction models (ARIMA, ETS, THETA) and three reconciliation strategies (BU, SHR, MSE) as traditional baselines.
    \item Deep learning baseline is the combination of the three state-of-the-art prediction models with three reconciliation methods:
    Prediction models:
    \begin{itemize}
        \item DeepAR: an approach to produce accurate probabilistic forecasts, based on autoregressive recurrent neural network model, which is widely applied in industrial forecasting tasks~\cite{salinas2020deepar}.
        
        \item NBEATS: deep neural architecture based on backward and forward residual links and a deep stack of neural layers to achieve interpretability \cite{oreshkin2019n}.
        \item Autoformer: we choose a representative former-based model, and a novel decomposition architecture with an Auto-Correlation mechanism for long term prediction~\cite{wu2021autoformer}.
    \end{itemize}
    Reconciliation Methods:
    \begin{itemize}
        \item BU: traditionally popular reconciliation method, with aggregation from bottom level forecast to maintain coherence~\cite{gross1990disaggregation}.
        \item Proj: the state-of- the-art end-to-end reconciliation method, with the closed-form projection minimizing the revision of base forecast while ensuring coherence. It is combined with DeepVar forecasting to form the best model in hierarchical forecasting in HierE2E~\cite{rangapuram2021end}.
        \item TDProb: an attention-based model to predict the distribution of allocation proportions from the parent node, and then apply top down allocate mechanism to ensure coherence~\cite{das2022deep}.
    \end{itemize}
    
    \item Four popular multivariate hierarchical time series forecasting methods: 
    \begin{itemize}
        \item HierE2E: an end-to-end framework for multivariate HTS that produce the coherent probabilistic forecasts, which applies closed-form projection reconciliation to ensure coherence~\cite{rangapuram2021end}.
        \item COPDeepAR: an end-to-end framework for temporal HTS that produce the coherent probabilistic forecasts for different granularity, which employs GNNs to modeling the temporal hierarchical structure and closed form projection reconciliation to ensure coherence~\cite{rangapuram2023, salinas2019high}.
        \item SHARQ: a flexible nonlinear model for multivariate HTS, which is trained with quantile losses and coherent penalty loss to ensure the coherence \cite{han2021simultaneously}.
        \item SLOTH: an end-to-end framework for multivariate HTS, which utilizes the hierarchical structure to produce the base forecasts and apply OptNet to ensure coherence~\cite{SLOTH_2023}.
    \end{itemize}
\end{itemize}

\section{Appendix D: Metrics Result for each Level of Experiment}
\label{app: level_result}
We present the MAPE of each level for all methods on three datasets for the forecasting experiment in Table \ref{tab:level_result}. 

\newcolumntype{s}{>{\hsize=12.0\hsize\scriptsize}X}
\setlength{\tabcolsep}{2pt}
\begin{table*}[htbp!]
    \centering
    \scriptsize
    \resizebox{\textwidth}{!}
    {
    \begin{tabularx}{0.95\textwidth}{s|ccc|ccc|cc}
        \hline
         Dataset & \multicolumn{3}{c}{Electricity}  & \multicolumn{3}{c}{Traffic} & \multicolumn{2}{c}{Exchange Rate}  \\ \hline
         Level Number & 1 & 2 & 3 & 1 & 2 & 3 & 1 & 2 \\ \hline
         THIEF-ARIMA-BU &0.0821 & 0.1240 & 0.1895 & 0.3316 & 0.5737 &  0.7501 & 0.0088 & 0.0095\\ 
         THIEF-ETS-BU  & 0.0723 &0.1382 &  0.2208  & 0.7034 &  0.9952 &0.8815 & 0.0073 &0.0100  \\
         THIEF-THETA-BU  & 0.0887 &  0.1344 & 0.2071 & 0.6326 & 0.7806 & 0.7064 & 0.0096 & 0.0102 \\
         THIEF-ARIMA-OLS  & 0.0813 & 0.1247& 0.1896 & 0.3141 & 0.5699 & 0.8293 & 0.0108 &  0.0118 \\
         THIEF-ETS-OLS  & 0.0723 & 0.1382 & 0.2208 & 0.3209 & 0.6928 & 1.0477 & 0.0098 & 0.0133 \\ 
         THIEF-THETA-OLS  &0.0818 & 0.1486 & 0.2256& 0.2565 & 0.7405 & 1.6886 & 0.0096 & 0.0094 \\
         THIEF-ARIMA-MSE  & 0.0835 & 0.1257  & 0.1906 & 0.3095 & 0.5327 & 0.7000 & 0.0089 &0.0097  \\
         THIEF-ETS-MSE  &0.0883 &  0.1395 & 0.2142 & 0.5157 & 0.8143& 0.7315 & 0.0095 & 0.0103  \\
         THIEF-THETA-MSE  & 0.0775 & 0.1308 & 0.2036 &  0.4468 &  0.7171 & 0.9467 & 0.0095 & 0.0094\\ \hline
         DeepAR-BU & 0.4548(0.0082) & 0.3912(0.0230) &0.4167(0.0181) & 0.6447(0.0440) & 0.6768(0.0320) &  1.3439(0.0171) & 0.0133(0.0030) & 0.0138(0.0029)\\ 
         NBeats-BU & 0.3821(0.0186) & 0.2495(0.0420)& 0.2683(0.0382)&0.3641(0.0062)& 0.3633(0.0438)& 0.5601(0.0569) & 0.0187(0.0065) & 0.0195(0.0062)\\ 
         Autoformer-BU & 0.4429(0.0787) & 0.3657(0.0978) & 0.4059(0.1021) & 0.4955(0.0252)&0.6760(0.1100) &1.6798(0.2757) & 0.0082(0.0006) & 0.0092(0.0006)\\
         DeepAR-Proj & 0.3889(0.0027) & 0.3610(0.0054) & 0.4166(0.0042) & 0.4039(0.0049) & 1.0373(0.0087) & 3.3717(0.0289) & 0.0107(0.0020) & 0.0114(0.0018)\\ 
         NBeats-Proj & 0.3785(0.0292)& 0.2180(0.0170) & 0.2432(0.0173) & 0.4138(0.0113) &0.4363(0.0140) & 0.5308(0.0676) & 0.0211(0.0072) & 0.0207(0.0075)\\ 
         Autoformer-Proj &0.4057(0.0123) & 0.2260(0.0064) & 0.2753(0.0047)& 0.4221(0.0105) & 0.4707(0.0493) & 1.1376(0.0757) & 0.0102(0.0007) & 0.0111(0.0006)\\ 
         DeepAR-TDProb & 0.3760(0.0010) & 0.1588(0.0047) & 0.1760(0.0055) &  0.3942(0.0012) & 0.4886(0.0113) & 1.0536 (0.0296) & 0.0090(0.0007) &0.0096(0.0008)\\ 
         NBeats-TDProb &0.3643(0.0125) & 0.1712(0.0108) & 0.1829(0.0081)&0.4132(0.0129) &0.5109(0.0138) & 1.0744 (0.00339) & 0.0157 (0.0062) & 0.0160(0.0060)\\ 
         Autoformer-TDProb & 0.4123(0.0297) & 0.1935(0.0265) & 0.2051(0.0242)& 0.4028(0.0107) & 0.4809(0.0163) &0.9894(0.0394)  & 0.0115 (0.0035)& 0.0120 (0.0034)\\ \hline
         HierE2E & 0.4267(0.0101) & 0.2326(0.0168)& 0.2346(0.0147) & 0.5068(0.0103) & 0.4508(0.0133) & 0.4728(0.0179) & 0.0125 (0.016)& 0.0145(0.0017)\\ 
         SHARQ & 0.4045(0.0065) & 0.2044(0.0057) &0.2101(0.0058) & 0.5283(0.0149) & 0.4799(0.0161)& 0.5041(0.0085) & 0.0120(0.0023) & 0.0156(0.0042)\\ 
         COPDeepAR &  0.9742(0.0580) & 0.6097(0.0517) & 0.6087(0.0530)& 0.8245(0.0068) & 0.8017(0.0063)& 0.8344(0.0058) &0.1244(0.0276) & 0.1247(0.0273)\\ 
         SLOTH &  0.3741(0.0036) & 0.1662(0.0024) & 0.1765(0.0021)& 0.4786(0.0013) & 0.4441(0.0020)& 0.4620(0.0025) &0.0106(0.0002) & 0.0113(0.0002)\\ 
         \hline
         GMP-BU &0.3573(0.0070) & 0.1578(0.0099) & 0.1710(0.0073) & 0.3782(0.0121)& 0.3530(0.0133)& 0.4484(0.0210) & 0.0097(0.0015) & 0.0106(0.0014) \\ 
         GMP-TDProb & 0.3710(0.0082) & 0.1552(0.0027) & 0.1731(0.0264) & 0.3626(0.0029)& 0.4525(0.0053)& 0.9989(0.0300) & 0.0082(0.0009) & 0.0088(0.0009)\\ 
         GMP-Proj &0.3722(0.0126) & 0.1783(0.0140) & 0.1936(0.0142) & 0.3877(0.0036)  &0.3504(0.0071) & 0.4867(0.0243) &  0.0083(0.0025) & 0.0086(0.0024)\\ 
         GMP-AR & 0.3606(0.0067) & 0.1368(0.0020) & 0.1499(0.0020) & 0.3736(0.0070) & 0.3368(0.0114) & 0.4288(0.0201) & 0.0078(0.0021) & 0.0084(0.0020)\\ \hline
    \end{tabularx}
    }
    \caption{MAPE values for each level over five independent runs for baselines such as traditional reconciliation methods, deep learning methods, and popular multivariate HTS methods,  as well as our approach. The values in brackets is the variances over five runs.}
    \label{tab:level_result}
\end{table*}

\section{Appendix E: Time Evaluation}
\label{app:time_eval}
We report the running time of each independent run for each method in this section in Table~\ref{tab:running_time}.
\begin{table}[t]
    \centering
    \begin{tabular}{|l|c|c|c|}
        \hline
         \textbf{Deep Model}  & Electricity & Traffic  & Exchange Rate \\
         \hline
         DeepAR-BU & 3045.12 & 6548.93 & 311.18 \\
         Nbeats-BU &  1434.85 &  2710.14& 502.01  \\
         Autoformer-BU & 3306.81 & 6442.07 & 602.87 \\
         DeepAR-Proj & 3103.58  & 6666.99 & 323.17  \\
         Nbeats-Proj & 1492.18 & 2460.88 &  588.58 \\
         Autoformer-Proj & 3325.0 & 6763.61 & 656.39 \\
         DeepAR-TDProb & 845.14  & 1531.86 &236.09\\
         Nbeats-TDProb &  983.7&  1673.52& 378.36 \\
         Autoformer-TDProb & 1242.08 & 2331.4 & 398.62 \\
         \hline
         HierE2E&  908.83 & 1606.43 & 230.02 \\
         SHARQ & 914.18 & 1566.39 & 224.83 \\
         COPDeepAR & 1079.15 & 1957.78  & 267.16 \\
         SLOTH & 1911.39 & 2539.19  & 374.88 \\
         \hline
         GMP-BU & 1026.05 & 1874.63 & 240.47 \\
         GMP-Proj & 1078.5 & 2098.05 & 258.84 \\
         GMP-TDProb &  1073.23& 1996.94 & 255.11  \\
         GMP-AR & 1075.02 & 2016.55 & 252.73 \\
         \hline
    \end{tabular}
    \caption{The average running time (seconds) over five independent runs for baselines such as traditional reconciliation methods and end-to-end methods, as well as our approach. }
    \label{tab:running_time}
\end{table}

\section{Appendix F: Ablation Study}

\subsection{Child Feature Fusion Strategy}
\label{app:ablation_chidren}
In this section, we compare our CNN strategy with no fusion or Attention fusion mechanism on the child's proportion modeling module on three publicly available datasets.
\begin{figure}[t]
    \centering
    \includegraphics[width=0.8\linewidth]{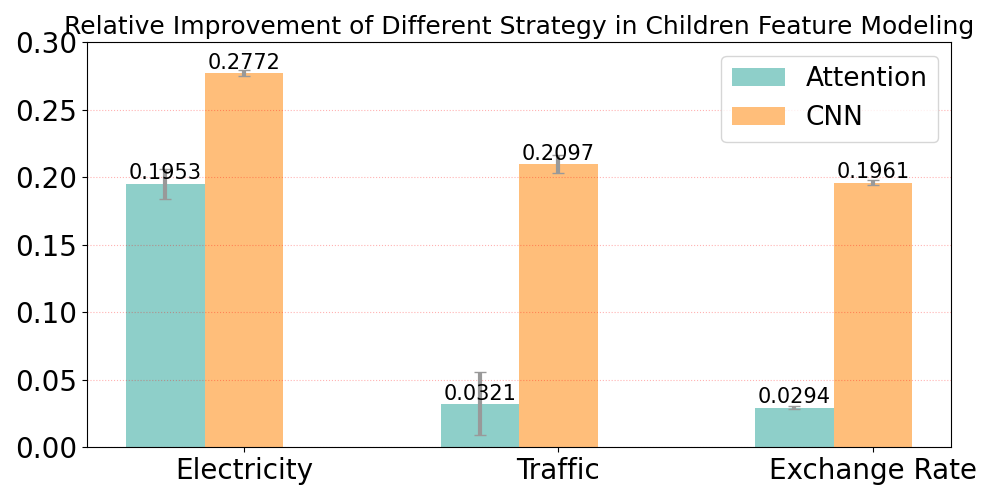}
    \caption{The figure shows relative improvement of five independent runs of \textit{Attention} and our \textit{CNN} feature fusion mechanism compared to utilizing temporal feature directly without feature fusion. }   \label{fig:ablation_children}
    \vspace{-5mm}
\end{figure}
Results in Fig.~\ref{fig:ablation_children} show that both Attention and CNN fusion mechanisms perform better compared to direct forecast on all datasets by 22.76\% and 8.55\%. Besides, CNN improves by nearly 14.21\% on average compared to Attention.

\subsection{Forecasting Performance Analysis for Component in Granularity Input Transformation}
\label{app:ablation_component}
\begin{figure}[t]
    \centering
    \includegraphics[width=0.8\linewidth]{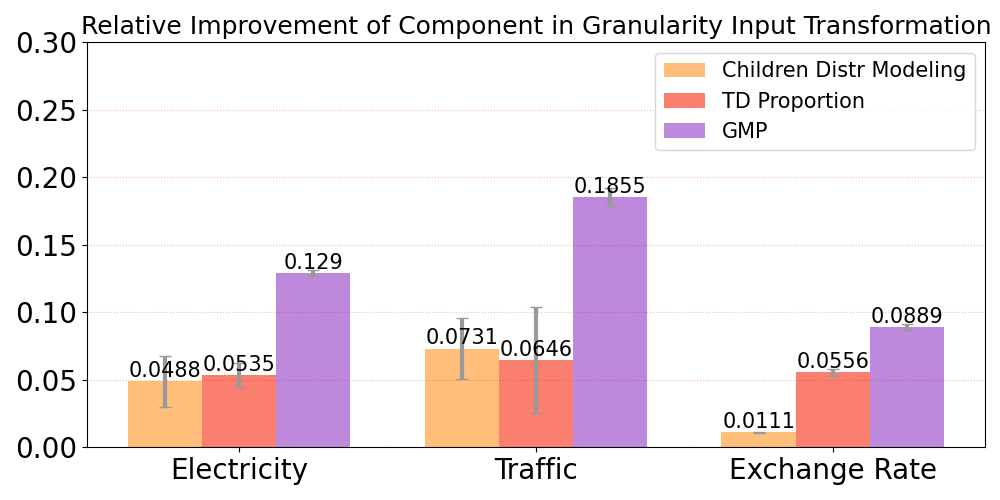}
    \caption{The figure shows relative improvement of five independent runs of \textit{Child Distribution Modeling}, \textit{Top Down Proportion} and the combination of them (GMP) input transformation mechanism compared to utilizing vanilla time series inputs. }   \label{fig:ablation_component}
    \vspace{-5mm}
\end{figure}

In this section, we evaluate the improvement of each component in granularity input transformation compared to vanilla temporal input.
As Fig.~\ref{fig:ablation_component} shows, children distribution modeling and top-down proportion (TD Proportion) can improve the performance relatively by 4.34\% and 5.79\%. And the combination of them can reach a 13.47\% improvement compared to vanilla input.

\subsection{Comparsion between Fixed Weight and Neural Weight}
\label{app:ablation_fixweight}
In this section, we compared the neural weight projection to 5 different types of fixed weights (C1-C5), as well as no weight on three datasets. The five different types of weights are formed as follows:
\begin{itemize}
    \item Root Node Weight (C1): we set the weight of the root node double as others in this setting.
    \item Middle Node Weight (C2): we set the weight of the nodes at the middle level double as others in this setting.
    \item Leaf Node Weight (C3): we set the weight of nodes at the bottom level as double as others in this setting.
    \item Random Weight (C4): we set the weight of all nodes randomly.
    \item Past Value Weight (C5): we set the weights of all nodes depending on their past values, including the means and standard values.
\end{itemize}

Results in Table \ref{tab:result_fix} show that reasonable fixed weight projection can improve the forecasting performance compared to model with no weight. Specifically, in our settings,  nearly 2/5 settings perform better than that of no weight. However, both all fixed weights and no weight are inferior to our neural weight method.

\newcolumntype{s}{>{\hsize=10\hsize}X}
\setlength{\tabcolsep}{3pt}
\begin{table*}[h]
    \centering
    \begin{tabularx}{\textwidth}{s|cc|cc|cc}
        \hline
         \textbf{Dataset} & \multicolumn{2}{c}{Electricity}  & \multicolumn{2}{c}{Traffic} & \multicolumn{2}{c}{Exchange Rate}  \\ \hline
         \textbf{Metrics} & b-MAPE & MAPE & b-MAPE& MAPE & b-MAPE & MAPE \\ 
         \hline
         GMP-Proj &0.1937(0.0143) & 0.2305(0.0068) & 0.4868(0.0243) & 0.4140(0.0145)  &0.0089(0.0024) & 0.0089(0.0018) \\ 
         GMP-Proj-C1 &0.1781(0.0065) & 0.2438(0.0125) & 0.4940(0.0309) & 0.4079(0.0183)  &- & -\\ 
         GMP-Proj-C2 &0.1888(0.0119) & 0.2330(0.0123) & 0.4914(0.0432) & 0.4038(0.0085)  &0.0089(0.0008) & 0.0095(0.0014) \\ 
         GMP-Proj-C3 &0.1820(0.0126) & 0.2481(0.0131) & 0.4772(0.0188) & 0.4083(0.0094)  &0.0103(0.0014) & 0.0086(0.0025) \\ 
         GMP-Proj-C4 &0.1862(0.0199) & 0.2360(0.0117) & 0.4844(0.0162) & 0.4096(0.0074)  &0.0089(0.0008) & 0.0085(0.0008) \\ 
         GMP-Proj-C5 &0.1750(0.0039) & 0.2299(0.0046) & 0.6234(0.0456) & 0.4656(0.0232)  &0.0090(0.0016) & 0.0086(0.0017) \\ 
         GMPAR & \textbf{0.1499(0.0021)} & \textbf{0.2158(0.003)} & \textbf{0.4289(0.0202)}& \textbf{0.3798(0.0066)} &\textbf{0.0085(0.0021)} & \textbf{0.0082(0.0021)}\\ \hline
    \end{tabularx}
    \caption{B-MAPE (bottom MAPE) and MAPE metric values over five independent runs for weight projection reconciliation such as naive projection, five different fixed weights projection, and our neural weighted reconciliation approach. The values in brackets are the variances over five runs.}
    \label{tab:result_fix}
    \vspace{-5mm}
\end{table*}


\section{Appendix G: Experiment Setup}
\label{app:exeriment_setup}
\paragraph{Computation Cost}
We conduct all our experiments on a server with the configurations of 2 NVIDIA Tesla P100-PCIe GPUs (16GB memory), 4 Intel(R) Xeon(R) CPU E5-2682 v4 @ 2.50GHz(64 cores) CPUs,  256GB memory and 400TB disk. The running time for each deep learning method is shown in Appendix E.

\paragraph{Hyperpameters Setting}
For all RNN-related neural methods,  the context length of the recurrent module is set in the range \{1, 3, 4, 8\} for all datasets. For deep learning methods, the number of hidden layers is chosen from range \{32, 64, 128\}. 
For Autoformer, the layer of encoder
is chosen from range \{2, 3, 4\}, the head number of multi-head attention is set as 8, the kernel size is set as 25, and the dimension of attention is set as 512. For NBEATS, the number of stacks is chosen from range \{32, 48, 64\}, the number of hidden state of fully connect layer is chosen from \{32, 64, 128\}, and the number of blocks is chosen from range \{1, 2, 3\}, and the number of block layers is chosen from range \{2, 4\}. 
For statistical methods, we follow the setting of Thief package directly. And the implementation and other details can check in the implementation code in the supplementary files.

\section{ongoing works}
In future works, we can induce the invariance learning in continuous domains to avoid OOD for test time range \citep{lin2023continuous}, and also can induce graph knowledge to hierarchical base model to improve performance\cite{yang2020relational, xie2021embedding}. Then try deep optimal ways to improve performance\cite{pan2023deep}. We can also do a system benchmark of temporal hts\cite{xue2023easytpp}.


%% file: aaai2024.bbl
\begin{thebibliography}{31}
\providecommand{\natexlab}[1]{#1}

\bibitem[{Athanasopoulos, Ahmed, and
  Hyndman(2009)}]{athanasopoulos2009hierarchical}
Athanasopoulos, G.; Ahmed, R.~A.; and Hyndman, R.~J. 2009.
\newblock Hierarchical forecasts for Australian domestic tourism.
\newblock \emph{International Journal of Forecasting}, 25(1): 146--166.

\bibitem[{Athanasopoulos et~al.(2017)Athanasopoulos, Hyndman, Kourentzes, and
  Petropoulos}]{thts_2017}
Athanasopoulos, G.; Hyndman, R.~J.; Kourentzes, N.; and Petropoulos, F. 2017.
\newblock Forecasting with temporal hierarchies.
\newblock \emph{European Journal of Operational Research}, 262(1): 60--74.

\bibitem[{Bai, Kolter, and Koltun(2018)}]{bai2018empirical}
Bai, S.; Kolter, J.~Z.; and Koltun, V. 2018.
\newblock An empirical evaluation of generic convolutional and recurrent
  networks for sequence modeling.
\newblock arXiv:1803.01271.

\bibitem[{Chen, Ma, and Lin(2021)}]{multiscale_rnn21}
Chen, Z.; Ma, Q.; and Lin, Z. 2021.
\newblock Time-Aware Multi-Scale RNNs for Time Series Modeling.
\newblock In Zhou, Z.-H., ed., \emph{International Joint Conference on
  Artificial Intelligence, {IJCAI-21}}, 2285--2291.

\bibitem[{Chung et~al.(2014)Chung, Gulcehre, Cho, and
  Bengio}]{chung2014empirical}
Chung, J.; Gulcehre, C.; Cho, K.; and Bengio, Y. 2014.
\newblock Empirical evaluation of gated recurrent neural networks on sequence
  modeling.
\newblock \emph{arXiv preprint arXiv:1412.3555}.

\bibitem[{Das et~al.(2022)Das, Kong, Paria, and Sen}]{das2022deep}
Das, A.; Kong, W.; Paria, B.; and Sen, R. 2022.
\newblock A deep top-down approach to hierarchically coherent probabilistic
  forecasting.

\bibitem[{Gardner and Dorling(1998)}]{gardner1998artificial}
Gardner, M.~W.; and Dorling, S. 1998.
\newblock Artificial neural networks (the multilayer perceptron)—a review of
  applications in the atmospheric sciences.
\newblock \emph{Atmospheric environment}, 32(14-15): 2627--2636.

\bibitem[{Gross and Sohl(1990)}]{gross1990disaggregation}
Gross, C.~W.; and Sohl, J.~E. 1990.
\newblock Disaggregation methods to expedite product line forecasting.
\newblock \emph{Journal of forecasting}, 9(3): 233--254.

\bibitem[{Han, Dasgupta, and Ghosh(2021)}]{han2021simultaneously}
Han, X.; Dasgupta, S.; and Ghosh, J. 2021.
\newblock Simultaneously Reconciled Quantile Forecasting of Hierarchically
  Related Time Series.
\newblock In \emph{International Conference on Artificial Intelligence and
  Statistics}, 190--198. PMLR.

\bibitem[{Kourentzes and Athanasopoulos(2019)}]{Kourentzes2019Cross}
Kourentzes, N.; and Athanasopoulos, G. 2019.
\newblock Cross-temporal coherent forecasts for Australian tourism.
\newblock \emph{Annals of Tourism Research}, 75.

\bibitem[{Laptev et~al.(2017)Laptev, Yosinski, Li, and Smyl}]{traffic2017}
Laptev, N.; Yosinski, J.; Li, L.~E.; and Smyl, S. 2017.
\newblock Time-series Extreme Event Forecasting with Neural Networks at Uber.
\newblock In \emph{International Conference on Machine Learning}, 1--5. PMLR.

\bibitem[{Li et~al.(2018)Li, Yu, Shahabi, and Liu}]{li2018diffusion}
Li, Y.; Yu, R.; Shahabi, C.; and Liu, Y. 2018.
\newblock Diffusion Convolutional Recurrent Neural Network: Data-Driven Traffic
  Forecasting.
\newblock In \emph{International Conference on Learning Representations}.

\bibitem[{Lin et~al.(2023)Lin, Zhou, Tan, Ma, Liu, He, Yuan, Liu, Zhang, Yang
  et~al.}]{lin2023continuous}
Lin, Y.; Zhou, F.; Tan, L.; Ma, L.; Liu, J.; He, Y.; Yuan, Y.; Liu, Y.; Zhang,
  J.; Yang, Y.; et~al. 2023.
\newblock Continuous Invariance Learning.
\newblock \emph{arXiv preprint arXiv:2310.05348}.

\bibitem[{Nocedal and Wright(2006)}]{num_optim}
Nocedal, J.; and Wright, S.~J. 2006.
\newblock \emph{Numerical Optimization}.
\newblock Springer New Yrok, NY.

\bibitem[{Oreshkin et~al.(2019)Oreshkin, Carpov, Chapados, and
  Bengio}]{oreshkin2019n}
Oreshkin, B.~N.; Carpov, D.; Chapados, N.; and Bengio, Y. 2019.
\newblock N-BEATS: Neural basis expansion analysis for interpretable time
  series forecasting.
\newblock arXiv:1905.10437.

\bibitem[{Pan et~al.(2023)Pan, Zhou, Hu, Zhu, Ning, Zhuang, Xue, Zhang, and
  Hu}]{pan2023deep}
Pan, C.; Zhou, F.; Hu, X.; Zhu, X.; Ning, W.; Zhuang, Z.; Xue, S.; Zhang, J.;
  and Hu, Y. 2023.
\newblock Deep optimal timing strategies for time series.
\newblock \emph{arXiv preprint arXiv:2310.05479}.

\bibitem[{Rangapuram et~al.(2023)Rangapuram, Kapoor, Nirwan, Mercado,
  Januschowski, Wang, and Bohlke-Schneider}]{rangapuram2023}
Rangapuram, S.~S.; Kapoor, S.; Nirwan, R.~S.; Mercado, P.; Januschowski, T.;
  Wang, Y.; and Bohlke-Schneider, M. 2023.
\newblock Coherent Probabilistic Forecasting of Temporal Hierarchies.
\newblock In Ruiz, F.; Dy, J.; and van~de Meent, J.-W., eds.,
  \emph{International Conference on Artificial Intelligence and Statistics},
  volume 206 of \emph{Proceedings of Machine Learning Research}, 9362--9376.
  PMLR.

\bibitem[{Rangapuram et~al.(2021)Rangapuram, Werner, Benidis, Mercado,
  Gasthaus, and Januschowski}]{rangapuram2021end}
Rangapuram, S.~S.; Werner, L.~D.; Benidis, K.; Mercado, P.; Gasthaus, J.; and
  Januschowski, T. 2021.
\newblock End-to-End Learning of Coherent Probabilistic Forecasts for
  Hierarchical Time Series.
\newblock In \emph{International Conference on Machine Learning}, 8832--8843.
  PMLR.

\bibitem[{Rostami-Tabar et~al.(2013)Rostami-Tabar, Babai, Syntetos, and
  Ducq3}]{demandforecast13}
Rostami-Tabar, B.; Babai, M.~Z.; Syntetos, A.; and Ducq3, Y. 2013.
\newblock Demand Forecasting by Temporal Aggregation.
\newblock \emph{Naval Research Logistics}, 60(6): 479--498.

\bibitem[{Salinas et~al.(2019)Salinas, Bohlke-Schneider, Callot, Medico, and
  Gasthaus}]{salinas2019high}
Salinas, D.; Bohlke-Schneider, M.; Callot, L.; Medico, R.; and Gasthaus, J.
  2019.
\newblock High-dimensional multivariate forecasting with low-rank gaussian
  copula processes.
\newblock \emph{arXiv preprint arXiv:1910.03002}.

\bibitem[{Salinas et~al.(2020)Salinas, Flunkert, Gasthaus, and
  Januschowski}]{salinas2020deepar}
Salinas, D.; Flunkert, V.; Gasthaus, J.; and Januschowski, T. 2020.
\newblock DeepAR: Probabilistic forecasting with autoregressive recurrent
  networks.
\newblock \emph{International Journal of Forecasting}, 36(3): 1181--1191.

\bibitem[{Sutskever, Vinyals, and Le(2014)}]{sutskever2014sequence}
Sutskever, I.; Vinyals, O.; and Le, Q.~V. 2014.
\newblock Sequence to sequence learning with neural networks.
\newblock \emph{Advances in neural information processing systems}, 27.

\bibitem[{Taieb, Taylor, and Hyndman(2021)}]{taieb2021hierarchical}
Taieb, S.~B.; Taylor, J.~W.; and Hyndman, R.~J. 2021.
\newblock Hierarchical probabilistic forecasting of electricity demand with
  smart meter data.
\newblock \emph{Journal of the American Statistical Association}, 116(533):
  27--43.

\bibitem[{Theodosiou and Kourentzes(2021)}]{theodosiou2021forecasting}
Theodosiou, F.; and Kourentzes, N. 2021.
\newblock Forecasting with deep temporal hierarchies.
\newblock \emph{Available at SSRN 3918315}.

\bibitem[{Vaswani et~al.(2017)Vaswani, Shazeer, Parmar, Uszkoreit, Jones,
  Gomez, Kaiser, and Polosukhin}]{vaswani2017attention}
Vaswani, A.; Shazeer, N.; Parmar, N.; Uszkoreit, J.; Jones, L.; Gomez, A.~N.;
  Kaiser, {\L}.; and Polosukhin, I. 2017.
\newblock Attention is all you need.
\newblock In \emph{Advances in neural information processing systems},
  5998--6008.

\bibitem[{Wickramasuriya, Athanasopoulos, and
  Hyndman(2019)}]{wickramasuriya2019optimal}
Wickramasuriya, S.~L.; Athanasopoulos, G.; and Hyndman, R.~J. 2019.
\newblock Optimal forecast reconciliation for hierarchical and grouped time
  series through trace minimization.
\newblock \emph{Journal of the American Statistical Association}, 114(526):
  804--819.

\bibitem[{Wu et~al.(2021)Wu, Xu, Wang, and Long}]{wu2021autoformer}
Wu, H.; Xu, J.; Wang, J.; and Long, M. 2021.
\newblock Autoformer: Decomposition transformers with auto-correlation for
  long-term series forecasting.
\newblock \emph{Advances in Neural Information Processing Systems}, 34:
  22419--22430.

\bibitem[{Xie, Zhou, and Soh(2021)}]{xie2021embedding}
Xie, Y.; Zhou, F.; and Soh, H. 2021.
\newblock Embedding symbolic temporal knowledge into deep sequential models.
\newblock In \emph{2021 IEEE International Conference on Robotics and
  Automation (ICRA)}, 4267--4273. IEEE.

\bibitem[{Xue et~al.(2023)Xue, Shi, Chu, Wang, Zhou, Hao, Jiang, Pan, Xu, Zhang
  et~al.}]{xue2023easytpp}
Xue, S.; Shi, X.; Chu, Z.; Wang, Y.; Zhou, F.; Hao, H.; Jiang, C.; Pan, C.; Xu,
  Y.; Zhang, J.~Y.; et~al. 2023.
\newblock Easytpp: Towards open benchmarking the temporal point processes.
\newblock \emph{arXiv preprint arXiv:2307.08097}.

\bibitem[{Yang et~al.(2020)Yang, Chen, Zhou, Gao, and Cao}]{yang2020relational}
Yang, F.; Chen, L.; Zhou, F.; Gao, Y.; and Cao, W. 2020.
\newblock Relational state-space model for stochastic multi-object systems.
\newblock \emph{arXiv preprint arXiv:2001.04050}.

\bibitem[{Zhou et~al.(2023)Zhou, Pan, Ma, Liu, Wang, Zhang, Zhu, Hu, Hu, Zheng,
  Lei, and Yun}]{SLOTH_2023}
Zhou, F.; Pan, C.; Ma, L.; Liu, Y.; Wang, S.; Zhang, J.; Zhu, X.; Hu, X.; Hu,
  Y.; Zheng, Y.; Lei, L.; and Yun, H. 2023.
\newblock SLOTH: Structured Learning and Task-Based Optimization for Time
  Series Forecasting on Hierarchies.
\newblock In \emph{Proceedings of the AAAI Conference on Artificial
  Intelligence}, volume~37, 11417--11425.

\end{thebibliography}
